\documentclass[lettersize,journal]{IEEEtran}
\usepackage{amsmath,amsfonts}
\usepackage{algorithmic}
\usepackage{algorithm}
\usepackage{array}
\usepackage[caption=false,font=normalsize,labelfont=sf,textfont=sf]{subfig}
\usepackage{textcomp}
\usepackage{stfloats}
\usepackage{url}
\usepackage{verbatim}
\usepackage{graphicx}
\usepackage{cite}
\usepackage{ulem}
\usepackage{multirow}
\usepackage{enumitem}
\usepackage{scicite}

\IEEEoverridecommandlockouts
\begin{document}

\title{Constrained Natural Language Action Planning for Resilient Embodied Systems}

\author{
Grayson Byrd$^{*1,2}$,
Corban Rivera$^{1,2}$,
Bethany Kemp$^{1,2}$,
Meghan Booker$^{1}$,

Aurora Schmidt$^{1}$,
Celso M de Melo$^{3}$,
Lalithkumar Seenivasan$^2$,
Mathias Unberath$^2$

\small$^{1}$Johns Hopkins University Applied Physics Laboratory,
\small$^{2}$Johns Hopkins University,
\small$^{3}$DEVCOM Army Research Lab,
\small$^{*}$Corresponding Author 

}

\maketitle

\begin{abstract}

Replicating human-level intelligence in the execution of embodied tasks remains challenging due to the unconstrained nature of real-world environments. Novel use of large language models (LLMs) for task planning seeks to address the previously intractable state/action space of complex planning tasks, but hallucinations limit their reliability, and thus, viability beyond a research context. Additionally, the prompt engineering required to achieve adequate system performance lacks transparency, and thus, repeatability. In contrast to LLM planning, symbolic planning methods offer strong reliability and repeatability guarantees, but struggle to scale to the complexity and ambiguity of real-world tasks. We introduce a new robotic planning method that augments LLM planners with symbolic planning oversight to improve reliability and repeatability, and provide a transparent approach to defining hard constraints with considerably stronger clarity than traditional prompt engineering. Importantly, these augmentations preserve the reasoning capabilities of LLMs and retain impressive generalization in open-world environments. We demonstrate our approach in simulated and real-world environments. On the ALFWorld planning benchmark, our approach outperforms current state-of-the-art methods, achieving a near-perfect 99\% success rate. Deployment of our method to a real-world quadruped robot resulted in 100\% task success compared to 50\% and 30\% for pure LLM and symbolic planners across embodied pick and place tasks. Our approach presents an effective strategy to enhance the reliability, repeatability and transparency of LLM-based robot planners while retaining their key strengths: flexibility and generalizability to complex real-world environments. We hope that this work will contribute to the broad goal of building resilient embodied intelligent systems. 


\end{abstract}


\section{Summary}

By leveraging the strengths of both emerging Large Language
Model (LLM)-based and well-understood symbolic planning components, we present a novel, hybrid approach to high-level embodied planning that allows for explicit and rigid constraint definition while preserving the adaptability and common-sense reasoning of its LLM components to enable a highly adaptable, reliable, repeatable, and transparent planning solution for use in unconstrained open-world environments.

\section{Introduction}
\IEEEPARstart{E}{nabling} the reliable autonomy of embodied agents in complex and potentially unknown environments is a long-standing goal in robotics. Achieving this goal requires seamless interaction and understanding between a variety of system components including perception, control, navigation, and high-level planning. Recently, the presumed reasoning capabilities of large language models (LLMs) have inspired their use as high-level planners for embodied applications. The flexibility of LLM planners, powered by their ``common sense'' understanding, makes them particularly useful in unconstrained environments in which unanticipated scenarios must be dynamically adapted to. Additionally, LLM compute time scales predominately with plan length instead of the number of possible outcomes, avoiding the prohibitive combinatorial explosion suffered by symbolic planning methods in sufficiently complex environments.

Despite the clear benefits of LLM planners, there are significant concerns regarding their reliability and repeatability as they are shown to hallucinate and generate erroneous next-action predictions \cite{liu2023llmpempoweringlargelanguage, skreta2023errorsusefulpromptsinstruction} without clear patterns to these failure modes. Additionally, building an LLM planner typically requires extensive and thoughtful prompt engineering \cite{prompt_engineering_citation} to constrain the planner to produce responses that effectively consider state information, action capabilities, safety requirements, and designer intent. Prompt engineering presents significant challenges due to the difficulty in managing ambiguity in human language, balancing specificity and flexibility, achieving repeatability across multiple model responses, identifying and mitigating biases, and addressing a lack of cross-model portability (i.e. the same prompt elicits wildly different behaviors across different or updated language models with new weights) \cite{Geroimenko2025}. Currently, relying solely on prompt engineering is unsafe, not transparent, and lacks prudence with regard to future model releases. Furthermore, it is unclear to what extent it is reasonable to describe a robotic system's intricacies through natural language. Practical systems depend on rigid and explicit logic to operate \textit{reliably} and \textit{repeatably}, while natural language is inherently ambiguous. Many robotic systems are idiosyncratically designed, commonly leading to fundamental conflicts in understanding between a language model's bias towards more traditional systems and the unique and potentially unorthodox characteristics of the current system, i.e. common sense is not always the best approach when attempting to understand and operate a specialized system through ambiguous high-level actions. This phenomenon was aptly described by Edgar Dijkstra in 1979 when he said: ``I suspect that machines to be programmed in our native tongues-be it Dutch, English, American, French, German, or Swahili-are as damned difficult to make as they would be to use.''

\begin{figure*}[!t]
  \centering
  \includegraphics[width=\linewidth]{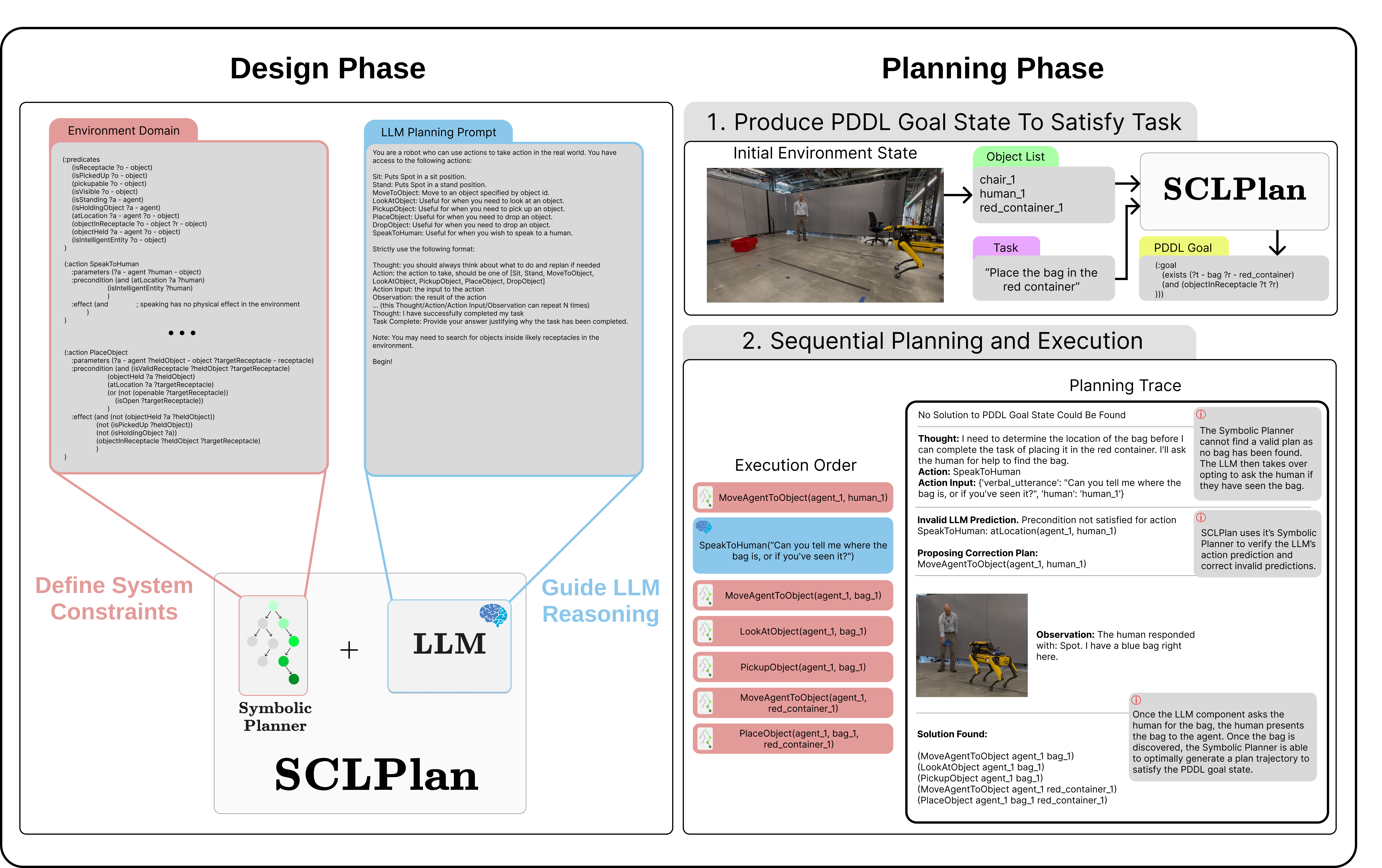}
  \caption{Overview of our \textbf{S}ymbolically \textbf{C}onstrained \textbf{L}anguage \textbf{P}lanner (SCLPlan). \textit{Design Phase}: In the design phase, an engineer can specify the planning constraints of their system (i.e. the preconditions and effects of each actions) through a PDDL \cite{pddl_original, pddl2} environment domain. This engineer can then create an LLM Planning Prompt that provides natural language descriptions of each of the available actions. \textit{Planning Phase}: During the planning phase, SCLPlan will first use an LLM to produce a PDDL goal state that can be used by its Symbolic Planner to generate a planning solution. Next, SCLPlan will sequentially plan to achieve the task, leveraging both LLM and Symbolic Planner components where necessary.}
  \label{fig:sclplan_overview}
\end{figure*}


In contrast to LLM planners, symbolic planning methods \cite{wilkins1984domain} achieve proven reliability by leveraging search algorithms to identify plans that satisfy a conjunctive goal state. Symbolic planning methods achieve this reliability through meticulously designed domains, which fully and rigidly define environmental constraints prior to planning. In accordance with Dijkstra's musings, the use of strict and precise logic makes symbolic planning methods ideal for describing specialized system constraints unambiguously, although this transparency comes at the expense of convenience and flexibility otherwise afforded by LLM planning approaches. Furthermore, explicitly defining a complete domain that \textit{fully} represents an environment is often infeasible for sufficiently complex environments or for unknown environments which cannot be wholly anticipated. Nonetheless, symbolic planners often provide optimality, reliability, and repeatability guarantees under sufficient levels of environmental certainty and understanding. While fully representing an entire open-world environment may be intractable, there are likely pockets of simplicity and certainty within such an environment that can be exploited by symbolic planning methods to solve simpler sub-tasks.

To fully realize the potential of LLM-based planning, methods must be developed to achieve the high levels of confidence and reliability required of real robotic systems while preserving strong adaptability in unconstrained environments. To this end, we present the \textbf{S}ymbolically \textbf{C}onstrained \textbf{L}anguage \textbf{Plan}ner (SCLPlan), a novel, hybrid approach to AI-based robot planning that addresses the challenges of hallucinations, prompt engineering, and ambiguity in natural language. An overview of SCLPlan along with an annotated planning example can be seen in Figure \ref{fig:sclplan_overview}. We develop a high-level AI planner which seamlessly blends an LLM with symbolic planning methods to create a framework in which key system constraints are strictly and transparently defined through principled logic while preserving strong adaptability under uncertainty. In moments of strong certainty within environments, we utilize a symbolic planner to optimally plan and execute tasks while relying on LLMs for planning needs that cannot be met symbolically, thus requiring stronger reasoning. Furthermore, we leverage the logical constraints defined for the symbolic planner to verify the output from our LLM planner, effectively catching and preventing the execution of hallucinated actions. Where prompt engineering lacks transparency, the introduction of formal logic into our language-based planner provides a principled way for engineers to constrain their systems, offering rigid and explicit constraints that meet the needs required of true real-world systems.

To assess the effectiveness of SCLPlan, we compare it to a baseline LLM planning method on the ALFWorld Planning Benchmark \cite{shridhar2020alfredbenchmarkinterpretinggrounded}, a text-based benchmark for natural language planning in household scenarios, and provide comprehensive ablation studies across several different LLM backbones. Additionally, we conduct experiments to determine the repeatability of our approach across various metrics during subsequent planning and execution rollouts of the same task to provide further insights into potential reliability improvements. We also create a new set of 16 difficult tasks in the AI2Thor simulator \cite{ai2thor} and provide a detailed analysis of the empirical success of SCLPlan in both environments, highlighting the utilization frequency and efficacy of various symbolic planner augmentations. Finally, we design and implement an exemplary real-world system on a quadruped robot with manipulator arm and create 10 accompanying real-world tasks. Tasks in this set are meant to be more complex and range from semantic pick and place tasks to tasks requiring ambiguous interaction with humans to complete the goal. We conduct similar evaluations in our real-world experiments. We find that our approach significantly improves performance across all experimental environments, representing a clear contribution to AI planning research. Most notably, we find that, averaged across 6 different LLM backbones, our approach achieves a Task Success rate improvement of +60\% compared to a Zero Shot baseline LLM planner on the ALFWorld validation set. Additionally, the repeatability of plan trajectories with regard to Task Success rate, Token Count, and Environment Steps is significantly improved when using our method with stronger LLM backbones. Detailed analyses from various ablation studies demonstrate that our planner effectively leverages it's symbolic planning methods in each environment and can effectively adapt its planning approach to the level of environmental complexity. Finally, we find that our approach transfers well to real-world environments, achieving 100\% success on our real-world environment and improving upon pure LLM and symbolic planners by +50\% and +70\%, respectively. These results were achieved with \textbf{no} change in the natural language prompt from the baseline LLM planner.

In summary, our contributions demonstrate that a combination of planning paradigms is indeed beneficial to planning performance when deployed in a variety of environments, both inside and outside simulation. The symbolic pieces of our approach provide a clear and logical process for defining hard constraints, circumventing the many issues and ambiguities associated with prompt optimization. We support these claims with rigorous quantitative and qualitative experiments. For the accompanying video, please refer to our Supplementary Materials. Our proposed method builds on several prior planning approaches; thus, we survey prior work serving as direct inspiration as well as related work that addresses similar open-world planning problems.

\textbf{Formal Task Planning Methods} - Robotic task planning focuses on devising a sequence of actions to accomplish predefined objectives within an environment. Conventional methods often rely on domain-specific languages (e.g., Planning Domain Definition Language (PDDL) \cite{pddl,pddl2}), finite state machines \cite{okrobot}, and techniques like semantic parsing and grammars \cite{parsing}, search algorithms \cite{search}, and heuristic-based optimization \cite{huristics} to generate solutions. However, these methods frequently encounter challenges with scalability and adaptability, particularly in high-dimensional, real-world environments where the state-action space grows exponentially. Alternatives such as hierarchical, imitation, and reinforcement learning-based approaches also struggle with data efficiency and scalability constraints \cite{rl,imitation}. To address these issues, our approach harnesses the common-sense reasoning abilities of Large Language Models (LLMs) to bypass the computationally expensive brute-force search problem inherent in task planning for open-world environments. 


\textbf{Task Planning with Large Language Models} - The use of LLMs for robotic task planning, particularly in translating natural language commands into executable task sequences, has been an area of increasing interest. Prior research has effectively utilized pre-trained LLMs’ contextual learning capabilities to generate actionable plans for embodied agents \cite{do_as_i_can,llmppddl,llmplanning,monologue,react,socratic,llmplanner,rivera2024conceptagent,booker2024embodiedrag, zhu2024knowagent}. However, LLM-induced hallucinations during sequential task planning are a constant threat to planning performance. Additionally, a persistent challenge for embodied planning is grounding generated task plans within the robot’s operational context and environmental state. Previous efforts have tackled this issue by integrating object detection models \cite{llmplanning}, PDDL environment representations \cite{llmppddl,sayplan}, or value function approximations \cite{do_as_i_can}. While these methods have shown promise, they provide only a loose guidance to the LLM as to what the robot's affordances are and do not provide reliable guarantees on the validity of the planning process.

\textbf{Retrieval-Augmented Generation for Task Planning} - Enhancing LLM reliability through the incorporation of external knowledge has emerged as a key strategy in improving task planning. This typically involves leveraging external tools to provide additional context or real-time feedback, guiding LLM-generated outputs. Prior work has explored various strategies for integrating external knowledge into LLM-based agents, including API calls to external services \cite{check,toolformer}, textual feedback from the operating environment \cite{llmsim,monologue,errors}, and domain randomization techniques \cite{tobin2017domainrandomizationtransferringdeep}. Building on these foundations, our framework introduces an agent-based approach that actively reacts and replans based on real-time observations. Unlike static task planning methods, our system dynamically refines its execution based on the evolving context, enabling it to complete natural language-specified tasks with greater adaptability.

\section{Results}

A summary of the results of the presented work can be seen in Movie 1. We develop a hybrid approach to AI planning and rigorously test performance on a variety of open-source and custom embodied task planning datasets in both simulated and real-world scenarios. Below, we provide an overview of our experiments and a detailed analysis of our results and findings.

\subsection{Experimental overview}

\begin{figure*}[htbp]
  \centering
  \includegraphics[width=\linewidth]{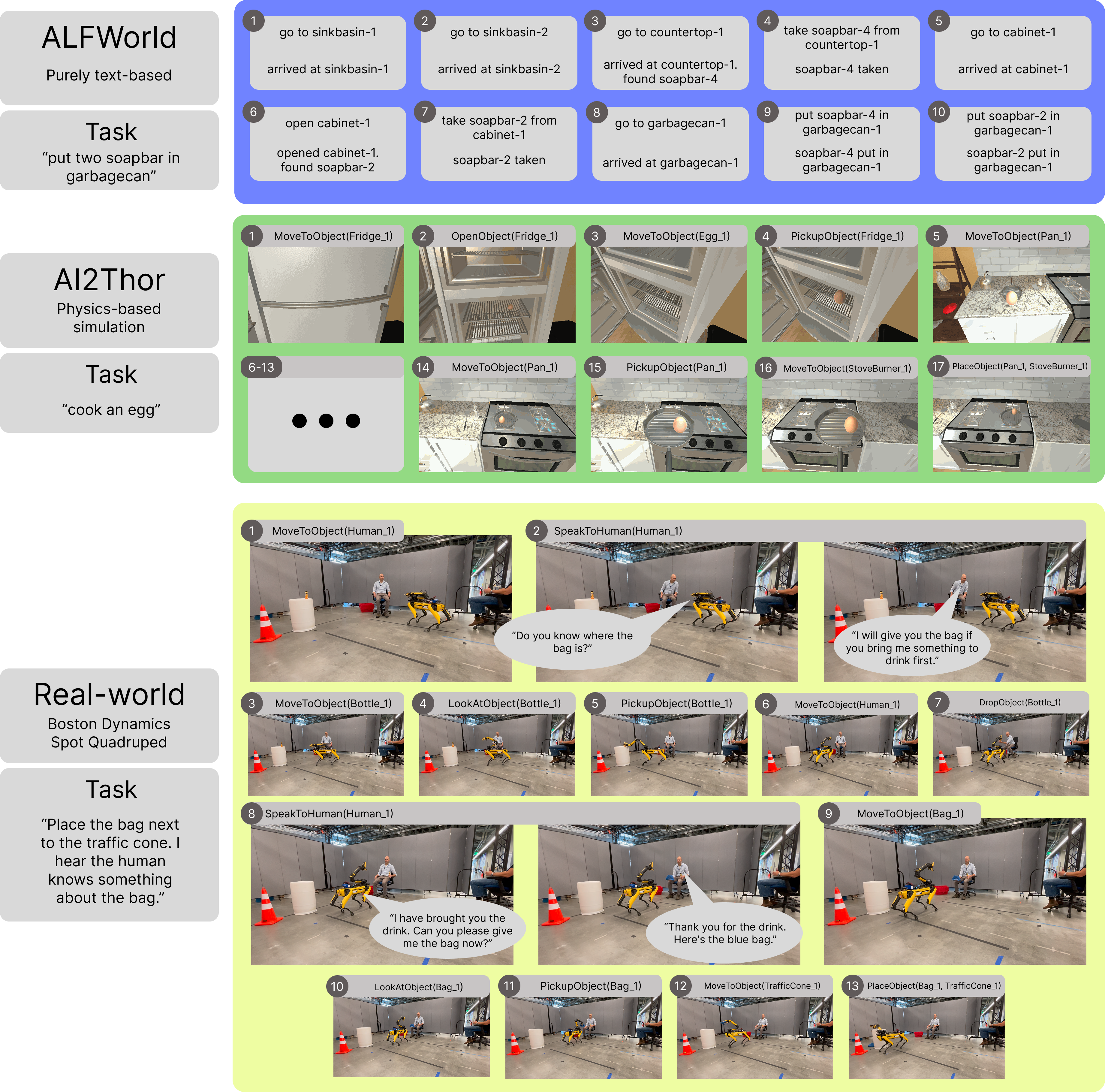}
  \caption{Task planning example from each experimental environment.}
  \label{fig:environment_overview}
\end{figure*}

We conduct our planning experiments across 3 different environments including a text-based, simulation-based, and real-world environment, as illustrated in Figure \ref{fig:environment_overview}. The use of text and simulation environments was chosen for two main reasons: to introduce diversity and scale in our benchmarking tasks and to provide a repeatable baseline to facilitate reliable and fair comparison to other approaches. We provide real-world experiments to demonstrate that our approach can be applied effectively to a real-world embodied system. Each environment contains its own unique corpus of planning problems as well as a unique set of actions available in that environment to the planning agent. Each action is a low-level policy representing a high-level `skill' acquired by the agent. The implementation details of each skill is irrelevant to the planning process, making our approach scalable and prudent with regards to accommodating new innovations in control as they are developed. Examples of possible actions found in our experiments include \textit{OpenObject}, \textit{MoveToObject}, and \textit{PickupObject}. Each agent also has access to a list of objects found in the environment that can be interacted with through the provided actions. This information is provided to the various LLM planners in the form of a natural language prompt where the list of actions remain static during a planning problem while the list of objects can dynamically change to accommodate newly discovered objects. To solve a planning problem, an agent is required to synthesize a coherent sequence of available actions to execute a high-level task provided in natural language. Example high-level tasks as well as valid task sequences to successfully complete them can be seen in Figure \ref{fig:environment_overview}.
We collect data across several metrics in addition to Task Success rate to get a richer understanding of the planning process, conduct ablation studies to illuminate the benefits of various design choices in our approach, and provide repeatability metrics to analyze the consistency of our approach across several rollouts of the same planning problem. Additionally, we conduct our experiments leveraging 7 different LLM backbones for our approach, ranging from expensive cloud models, such as GPT-4o, to small, locally run, open-source models such as Llama3.1-8b. For all experiments, no model training or finetuning is required -- we use strictly off-the-shelf models. For all ablation experiments, no LLM prompts were modified across ablations. Below, we provide a detailed account of each experimental environment.

\subsection{Description of Planning Environments}
\subsubsection{ALFWorld simulator and the ALFRED planning benchmark}

The ALFWorld \cite{shridhar2021alfworldaligningtextembodied} simulator provides an open-world, domestic environment for agents to navigate and interact with by initiating an interactive, parallel environment that aligns a textual description and action commands with an embodied robotic simulation (AI2Thor) \cite{ai2thor}. This provides a natural language interface well suited for language agents as the control of the physical robot in the simulation is already aligned with natural language commands. ALFWorld adopts tasks from the ALFRED dataset \cite{shridhar2020alfredbenchmarkinterpretinggrounded}, a benchmark for planning problems for embodied household tasks that uses natural language instructions. The ALFRED dataset consists of a large corpus of challenging embodied planning problems that require intelligent interaction with objects in the environment as well as navigation to and from various objects and locations in the environment. ALFWorld provides six different task types taken from the ALFRED dataset: pick and place, examine in light, clean and place, heat and place, cool and place, and pick two and place. Pick and place is the simplest task type, requiring an agent only to navigate to an object, pick it up, and place it elsewhere. All other task types require more complex reasoning, needing an agent to leverage multiple objects in the environment to appropriately change the state of other objects (e.g. using a lamp to examine a held object in the light). Each task provides a goal description annotation in natural language that describes what needs to be accomplished to complete the task successfully (e.g. ``put a bowl in the microwave''). To achieve sufficient variety among tasks in the ALFRED dataset, the embodied simulator contains 120 different rooms each falling into one of the following categories: kitchen, bedroom, bathroom, living room. Each room contains a set of randomly placed, moveable objects (e.g. bowl, pan) and static receptacles (e.g. microwave, table). Multiple possible tasks are provided for each room. To complete tasks in the embodied simulator, ALFWorld provides the following low-level action primitives in natural language for sequencing action trajectories: 

\begin{enumerate}
  \item \texttt{go\ to} \{recep\}
  \item \texttt{open} \{recep\}
  \item \texttt{clean} \{obj\} \texttt{with} \{recep\}
  \item \texttt{take}  \{obj\} \texttt{from} \{recep\}
  \item \texttt{close} \{recep\}
  \item \texttt{heat} \{obj\} \texttt{with} \{recep\}
  \item \texttt{put} \{obj\} \texttt{in/on} \{recep\}
  \item \texttt{toggle} \{obj\}/\{recep\}
  \item \texttt{cool} \{obj\} \texttt{with} \{recep\}
\end{enumerate}

ALFWorld provides a set of training and validation tasks. Since we do no training or finetuning of the models used in this work, we use only the validation set which consists of 134 unique and diverse tasks. We note that, while ALFWorld does contain parallel text-world and embodied simulation environments, use of the simulation environment has no effect on the task set and is used strictly for visualizing purposes. For this reason, we use only the text-world component of the ALFWorld environment.

\subsubsection{Custom AI2Thor Environment}

Although the ALFWorld simulator provides breadth and diversity in benchmark planning tasks for quality comparisons to other approaches and statistical rigor for ablation studies, we hypothesized that increased task complexity could elicit more frequent use of the LLM portion of our planner. Thus, we chose to create our own set of complex tasks in the AI2Thor simulator (used as the embodied sim component in ALFWorld as well) to better demonstrate how our planner performs on tasks that require more complex reasoning to accomplish. To do this, we provide a different set of action primitives to the agent that facilitate more complex reasoning. The action primitives for our custom AI2Thor experiments are: 

\begin{enumerate}
  \item \texttt{MoveToObject} \{obj\}/\{recep\}
  \item \texttt{OpenObject} \{recep\}
  \item \texttt{CloseObject} \{recep\}
  \item \texttt{PickupObject} \{obj\}
  \item \texttt{PlaceObject} \{obj\} in \{recep\}
  \item \texttt{SliceObject} \{obj\}
  \item \texttt{ToggleObjectOn} \{obj\}
  \item \texttt{ToggleObjectOff} \{obj\}
\end{enumerate}

Upon inspection, one might realize that there are less action primitives for our Custom AI2Thor Environment than the ALFWorld environment and intuitively assume that less action primitives would mean a less complex planning problem. The removal of various action primitives (i.e. heat, cool, clean) was intentional and actually increases the need for complex reasoning. These three action primitives represent effects that could be achieved implicitly through intelligent use of the environment (e.g. placing a bowl under running water in the sink to clean it or putting bread in the toaster and turning the toaster on to heat it up). To design a truly intelligent robot system, one should not have to explicitly program every possible action as this becomes intractable in an open-world environment. One should, however, be able to expect that an intelligent system could find ways, using what actions it does have at its disposal, to accomplish behaviors that are not explicitly defined. Thus, we created a new benchmark of 16 tasks in the AI2Thor environment that present more challenging planning scenarios in which an agent is expected to reason about how to achieve ambiguous tasks. Examples of these tasks include "Cook an egg.", "Rinse off the pan.", and "I have put my K-cup in the coffee machine. Now get a mug for me and make me coffee.". Since the agent has no action primitive to ``cook'', ``rinse'', or ``make coffee'', it must reason about how to leverage the actions it does possess to complete these example actions. We hypothesized that this increased ambiguity, coupled with longer-horizon planning sequences, would further highlight the benefits of our hybrid planning approach. It is worth noting that we would like to have added more than 16 tasks, but there are challenging scalability difficulties associated with increasing the number of tasks. This is because each task has multiple valid goal states that would satisfy the overall natural language command. For example, ``Cook the egg'' could be completed in many different ways. For this reason, each task planning and execution rollout had to be manually verified and scored by a human. As such, only 16 tasks were created and only two different LLM backbones (GPT-4o-mini and GPT-4o) were used for evaluation on these tasks.

\subsubsection{Spot Quadruped with Dexterous Manipulator Arm}

We further validate our approach on a real-world system using a Boston Dynamics Spot quadruped with dexterous manipulator arm as our robotic platform. We incorporate the necessary perception, localization, and navigation pipelines into our system from prior work. Details of these system components are provided in the Materials and Methods section. We test our approach extensively in the real-world by executing more than 40 planning rollouts on 10 unique tasks. In our real-world experiments, we provide the following action primitives to our robot:

\begin{enumerate}
  \item \texttt{Sit}
  \item \texttt{Stand}
  \item \texttt{MoveToObject} \{obj\}/\{recep\}
  \item \texttt{LookAtObject} \{obj\}/\{recep\}
  \item \texttt{PickupObject} \{obj\}
  \item \texttt{PlaceObject} \{obj\} in \{recep\}
  \item \texttt{DropObject}
  \item \texttt{SpeakToHuman} \{human\_name\}
\end{enumerate}

We choose our tasks carefully, ensuring that the types covered in our experiments cover a wide breadth of assignments that could be reasonably expected of a truly general embodied system. To highlight the benefits of a hybrid approach (and expose some of the shortcomings of isolated LLM and formal planners) we choose some tasks to be relatively repeatable, requiring minimal critical reasoning while benefiting from strong reliability guarantees, and also include dynamic tasks that necessitate strong reasoning capabilities and real-time adaptation. Task examples that are repeatable and require minimal reasoning include ``Place the apple and the bag in the red container.'' and ``Drop the bag by the traffic cone.'' These tasks are unambiguous but require a reliably sequenced task trajectory to complete, as any small planning error has the potential to derail the entire sequence. We hypothesized that purely LLM planners would struggle with such tasks while formal methods would perform quite reliably. An example of a complex and dynamic task presented to our agent is ``I want you to place a bag next to the traffic cone. I believe the human over there knows something about a bag.'' This required our agent to intelligently locate an unseen object (i.e. the bag) in the environment and interact with other intelligent entities to complete the task. The human then negotiated with the agent, stating that they would provide the bag to the agent if the agent first gave them something to drink. This task would be very difficult for a formal planner to complete alone, as it is impossible to predict the response from the human. Additionally, this task required natural language conversation, which is a reasonable expectation of a general and intelligent embodied agent. Although an LLM would likely be required without extensive tuning and engineering of a formal planner, we hypothesized the use of a formal planner to constrain and guide the output of the LLM portion of our planner could still prove beneficial for some portions of even the most ambiguous and dynamic tasks. A full description and discussion of our experimental results are shown below.

\subsection{Symbolic Planning Augmentations Improve Planning Performance in Simpler Environments}

\begin{figure*}[htbp]
  \centering
  \includegraphics[width=\linewidth]{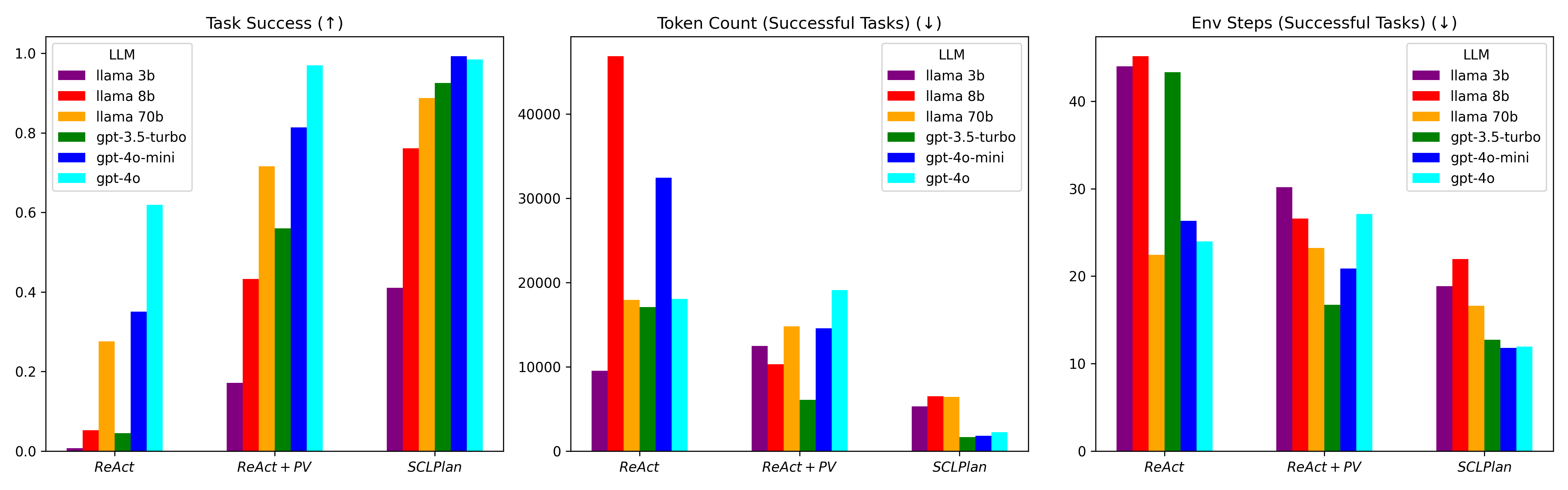}
  \caption{Ablation study of SCLPlan on ALFWorld task planning benchmark reveals significant increase in Task Success percentage, reduction in Token Count, and reduction in Environment Steps required to complete each task. These improvements persist across a variety of open source and cloud based LLMs of varying competence.}
  \label{fig:alfworld_general_ablation_study}
\end{figure*}

A core hypothesis of this work is that simple environments exist in which an LLM may be asked to complete a task in its entirety, but large portions of the task trajectory could be solved using a reliable symbolic planner. For example, while an LLM may be able to synthesize an action trajectory to pick up a bag and place that bag in a container, symbolic planners have been optimally planning for tasks of similar complexity for decades. So, why use an inherently unreliable LLM to synthesize the complete task trajectory if not entirely necessary? An LLM may still be required to \textit{complete} a task, e.g. strong semantic priors are needed to identify likely potential locations for the bag, but once the bag and container are found, a symbolic planner should be able to optimally solve the planning task. In this section, we discuss ALFWorld: an environment containing these types of simple planning problems, i.e. problems that may benefit from the use of LLM planners while not directly requiring them for the prediction of every single action in a viable action trajectory. The ALFWorld task set, while extensive, contains relatively simple tasks and available actions, resulting in lesser critical reasoning requirements than the other two experimental environments. As such, defining an effective domain for symbolic planning methods is a reasonable expectation. While a traditional symbolic planner alone cannot solve unstructured natural language tasks, we hypothesized that large portions of successful action sequences could be generated reliably from the comprehensive environment domain. To demonstrate that the use of formal method augmentations can significantly improve natural language task performance, we conduct extensive ablation studies spanning several different LLM backbones and evaluation metrics on the ALFWorld validation task dataset. As algorithms for robotic applications are often expected to run on the edge, potentially without internet access, we use both open source LLMs running locally and cloud-based models that are accessed via an online API. In Figure \ref{fig:alfworld_general_ablation_study}, we provide data for three different approaches: 1) $ReAct$\cite{react} - a pure LLM approach, 2) $ReAct + PV$ - a pure LLM whose next action predictions are first verified or corrected with our formal method Precondition Verification, and 3) $SCLPlan$ - our full AI planner, which includes Precondition Verification as well as a Global Symbolic Planner (GSP) that constantly searches for a global solution to the planning task at each next action prediction. We evaluate each approach using Task Success percentage, total Token Count, and number of Environment Steps required to complete each task as our experimental metrics.

Figure \ref{fig:alfworld_general_ablation_study} clearly demonstrates the benefits of each formal method augmentation on the ALFWorld validation task set. Across every LLM, regardless of size and competency, the introduction of Precondition Verification and then Global Formal Planning significantly improved the overall Task Success rate of the pipeline. This pattern generally extends to both Token Count and Environment Steps, with a notable exception for the formulation of the planner using Llama 70B as its language backbone, which experienced a slight degradation of performance on the Token Count and Environment Steps metrics when using $ReAct + PV$ alone. The improvements averaged across all LLMs are shown in Table \ref{tab:avg_ablation_improvements}. 

\begin{table}[htbp]
\setlength{\tabcolsep}{2pt}
\centering
\caption{Ablation study improvements averaged across each of the 7 language models tested.}
\begin{tabular}{lccc} \hline
\rule{0pt}{2.2ex}Method & Task Success (\%) & Token Count & Env Steps \\ \hline\hline \rule{0pt}{2ex}
$ReAct$& - & - & - \\
$ReAct + PV$& $+ 40$ & $- 10.8k$  & $-10.1$ \\
$ReAct + PV + GSP$& $+ 60$ & $- 19.7k$ & $-18.6$ \\\hline \rule{0pt}{2ex}
\end{tabular}
\label{tab:avg_ablation_improvements}
\end{table}

Overall, we see an average improvement of ($+40\%$) on Task Success and an average reduction in Token Count and Environment Steps of ($-10.8k$) and ($-10.1$) when adding Precondition Verification to our baseline $ReAct$ LLM planner. We see greater average improvements of ($+60\%$) on Task Success and a reduction in Token Count and Environment Steps of ($-19.7k$) and ($-18.6$) when incorporating the full implementation of our method complete with all augmentations. These results clearly demonstrate that our approach has the potential to significantly improve overall planning performance across a wide variety of metrics and language models. Additionally, our approach's strong performance when using weaker, local LLM backbones is particularly impactful as it shows potential in robotic applications where the use of smaller, local models is often required.

\subsection{Symbolic Planning Augmentations Improve Planning Repeatability in Simpler Environments}

Although the strong task performance across several metrics is indicative of $SCLPlan$'s proficiency, there are additional considerations when benchmarking intelligent task planning. One such consideration is repeatability. While there are often many different ways to successfully complete a sufficiently complex natural language task, extreme disparities in planning performance across subsequent planning rollouts of the same or similar task is undesirable for several reasons. On one hand, inconsistency in planning length makes it difficult to anticipate completion time, charging requirements for robots, and collaboration in multi-robot or human-robot teams. More seriously, a lack of repeatability in task completion can lead to lack of trust in a robotic system. Additionally, when using LLM planners, lack of repeatability in token count can make it difficult to accurately predict the cost associated with cloud-based models. LLMs have been shown to be inconsistent in their knowledge and output \cite{sosa2024reasoning}, thus we hypothesize that the inherent stochasticity of LLM planners hinder planning repeatability in both complex and simple tasks. For complex tasks where common-sense capabilities are paramount, this tradeoff between flexibility and repeatability may be warranted, but for simpler tasks or for simpler action sequences within a global planning trajectory, symbolic planners can be leveraged to provide increased reliability and repeatability. 

To explore the effect of each of our augmentations on planning repeatability, we perform an ablation study on the variation in the planning process. Specifically, we randomly sample 10 tasks from the 134 tasks in the ALFWorld validation task set. We then attempt to execute a plan using our approach on each task 10 times, keeping the environment consistent for each subsequent run of the same task. Next, we record the average and standard deviation values of each metric across the 10 runs of the same task and repeat this process for each of the 10 randomly sampled tasks. Finally, we ablate the various augmentations of our method and repeat the above process. The results, averaged across each of the 10 sampled tasks, are shown in Table \ref{tab:ablation_consistency}. Interestingly, when using less capable LLMs (i.e. Llama 3B and 8B), the repeatability tends to decrease as formal augmentations are added, although planning performance increases across all metrics. This suggests that while our augmentations do increase the chances of a weaker LLM to complete more difficult tasks, it does so intermittently causing a decrease in repeatability. For example, Llama 3B with a purely LLM approach very consistently fails at a rate of 97\%. Introducing our augmentations decreases that failure rate to 85\%, but doing so decreases the repeatability as it is a less extreme split between success and failure. Although this does decrease consistency, this is likely still desirable as it is improving overall performance. Stronger models (Llama 70B and larger), however, show significant improvements in repeatability across all metrics when compared to the purely LLM approach. 

\begin{table}
\setlength{\tabcolsep}{2pt}
\centering
\caption{Ablation study using repeatability metrics. The values presented represent the mean and standard deviation of each metric across 10 runs of the same task averaged across 10 different tasks. The lowest standard deviation for each model is \textbf{bolded}.}
\begin{tabular}{clccc} \hline
\rule{0pt}{2.2ex}LLM & Method & Task Success & Token Count & Env Steps \\ \hline\hline \rule{0pt}{2ex}\multirow{3}{*}{Llama 3B} & $ReAct$ & 0.03 (\textbf{±0.05}) & 42.8k (\textbf{±9.6k}) & 27.45 (\textbf{±4.26})\\
& $ReAct + PV$ & 0.15 (±0.16) & 34.6k (±13.9k) & 29.07 (±5.20)\\
& $SCLPlan$ & 0.32 (±0.32) & 26.2k (±15.1k) & 25.70 (±6.97)\\
\hline \rule{0pt}{2ex}
\multirow{3}{*}{Llama 8B} & $ReAct$ & 0.07 (\textbf{±0.09}) & 48.3k (±38.2k) & 31.07 (\textbf{±3.14})\\
& $ReAct + PV$ & 0.32 (±0.22) & 25.7k (\textbf{±12.7k}) & 31.34 (±5.82)\\
& $SCLPlan$ & 0.77 (±0.39) & 12.4k (±13.6k) & 23.58 (±11.45)\\
\hline \rule{0pt}{2ex}
\multirow{3}{*}{Llama 70B} & $ReAct$ & 0.36 (±0.24) & 26.9k (±7.2k) & 26.55 (±7.03)\\
& $ReAct + PV$ & 0.71 (±0.19) & 21.8k (±5.5k) & 28.89 (±10.19)\\
& $SCLPlan$ & 0.89 (\textbf{±0.13}) & 9.6k (\textbf{±4.6k}) & 18.03 (\textbf{±7.51})\\
\hline \rule{0pt}{2ex}
\multirow{3}{*}{GPT-4o-mini} & $ReAct$ & 0.43 (±0.39) & 33.9k (±16.2k) & 24.27 (±8.32)\\
& $ReAct + PV$ & 0.78 (±0.23) & 12.7k (±5.8k) & 18.28 (±7.50)\\
& $SCLPlan$ & 1.00 (\textbf{±0.00}) & 1.7k (\textbf{±0.1k}) & 11.13 (\textbf{±3.14})\\
\hline \rule{0pt}{2ex}
\multirow{3}{*}{GPT-4o} & $ReAct$ & 0.60 (±0.31) & 23.1k (±10.4k) & 24.54 (±6.73)\\
& $ReAct + PV$ & 0.94 (±0.10) & 15.5k (±5.1k) & 23.73 (±8.99)\\
& $SCLPlan$ & 1.00 (\textbf{±0.00}) & 3.8k (\textbf{±4.0k}) & 11.14 (\textbf{±4.52})\\\hline \rule{0pt}{2ex}
\end{tabular}
\label{tab:ablation_consistency}
\end{table}

These results show potential to greatly improve the reliability and transparency of AI planning approaches, as the agents will tend to act more predictably. The combination of significant overall performance improvements coupled with increased repeatability could have a profound impact on the trust between humans and intelligent embodied agents.

\subsection{Complex environments elicit increased LLM usage while still benefiting from a hybrid planning approach}

Another key insight from our experiments is that our hybrid planning approach can adapt its planning methodology to accommodate the level of complexity in the environment. Tables \ref{tab:alfworld_detailed_experiments} and \ref{tab:ai2thor_detailed_experiments} show a detailed analysis of the planning process on both the simpler ALFWorld and more complex AI2Thor experimental environments. These tables depict the results of detailed ablation experiments that provide insight into how SCLPlan uses its augmentations under different levels of environmental complexity. We provide ablations in the same form as previous experiments (adding Precondition Verification and then a Global Symbolic Planner to a baseline ReAct LLM planner) and show the proportion of actions generated by each component of the planner. We additionally provide data depicting the number of $ReAct$ predicted actions that were initially invalid followed by the percentage of these actions that were able to be successfully executed after using the Precondition Verification component. Finally, we also list the proportion of instances in which the Global Symbolic Planner was able to find a solution to the task at some point in the planning process. We provide these metrics when using both GPT-4o-mini and GPT-4o as LLM backbones.

\begin{table*}[htbp]
\centering
\caption{Detailed analysis of our method on the easier ALFWorld validation dataset.}
\begin{tabular}{clccc|ccc}
& & \multicolumn{3}{c}{GPT-4o-mini} & \multicolumn{3}{c}{GPT-4o} \\
&Metric& $ReAct$ & $ReAct + PV$ & $SCLPlan$ & $ReAct$ & $ReAct + PV$ & $SCLPlan$ \\ \hline \hline \rule{0pt}{2ex}
&Task Success (\textuparrow) &0.351 & 0.813 & 0.993 & 0.619 & 0.970 & 0.985\\
&Invalid $ReAct$ Predictions (\textdownarrow) & - & 0.262 & 0.006 & - & 0.233 & 0.032 \\
&Invalid $ReAct$ Predictions Corrected w/ $PV$ (\textuparrow) & - & 0.531 & 1.00 & - & 0.699 & 1.000 \\
& Global Symbolic Plan Found (\textuparrow) & - & - & 0.970 & - & - & 0.940 \\ \hline \rule{0pt}{2ex}
&Percentage $ReAct$ Actions& 1.000 & 0.804 & 0.188 & 1.000 & 0.789 & 0.213 \\
&Percentage $PV$ Actions & - & 0.196 & 0.002 & - & 0.211 & 0.013 \\
&Percentage $GSP$ Actions & - & - & 0.810 & - & - & 0.774 \\ \rule{0pt}{2ex}\\
\end{tabular}
\label{tab:alfworld_detailed_experiments}
\end{table*}

\begin{table*}[htbp]
\centering
\caption{Detailed analysis of our method on the more difficult Custom AI2Thor dataset.}
\begin{tabular}{clccc|ccc}
& & \multicolumn{3}{c}{GPT-4o-mini} & \multicolumn{3}{c}{GPT-4o} \\
&Metric& $ReAct$ & $ReAct + PV$ & $SCLPlan$ & $ReAct$ & $ReAct + PV$ & $SCLPlan$ \\ \hline \hline \rule{0pt}{2ex}
&Task Success (\textuparrow) &0.062 & 0.688 & 0.625 & 0.312 & 0.750 & 0.938\\
&Invalid $ReAct$ Predictions (\textdownarrow) & - & 0.370 & 0.369 & - & 0.295 & 0.261 \\
&Invalid $ReAct$ Predictions Corrected w/ $PV$ (\textuparrow) & - & 0.733 & 0.72 & - & 0.795 & 0.962 \\
& Global Symbolic Plan Found (\textuparrow) & - & - & 0.004 & - & - & 0.016 \\ \hline \rule{0pt}{2ex}
&Percentage $ReAct$ Actions & 1.000 & 0.657 & 0.635 & 1.000 & 0.675 & 0.646 \\
&Percentage $PV$ Actions & - & 0.343 & 0.349 & - & 0.325 & 0.290 \\
&Percentage $GSP$ Actions & - & - & 0.016 & - & - & 0.064 \\ \rule{0pt}{2ex}\\
\end{tabular}
\label{tab:ai2thor_detailed_experiments}
\end{table*}

The results clearly show that SCLPlan significantly improves planning performance compared to the $ReAct$ agent across each environment and LLM backbone, showing that SCLPlan's benefits extend to more complex environments. More interesting is the difference in planning strategies exhibited by SCLPlan in the different environments. In the simpler ALFWorld environment, the Global Symbolic Planner dominated the planning process; SCLPlan was consistently able to find global planning solutions to the task, doing so at an average rate of 97 \% and 94 \% for GPT-4o-mini and GPT-4o. Conversely, a global planning solution was found at a rate of only 0.4 \% and 1.6 \% on the more complex AI2Thor task set. This phenomenon is further shown in the proportion of actions predicted by each component in the planner as a staggering 77-81 \% of actions were predicted by the GSP for ALFWorld compared to only 1.6-6.4\% of actions for AI2Thor. Furthermore, the ReAct component contributed predictions at a much higher rate in the AI2Thor environment: 63.5-64.6\% compared to ALFWorld's 18.8-21.3\%. Unsurprisingly, the usage of the Precondition Verification component was used significantly more in the AI2Thor environment, likely due to the increased usage of the ReAct component, which presented more opportunities for the Precondition Verification to take effect. This supports the addition of Precondition Verification as it make up large swaths of the AI2Thor plan trajectories. One surprising insight is that the percentage of invalid ReAct predictions is similar for both environments except in the case of a simple environment using the full SCLPlan planner. This shows that even in simple environments, LLMs continue struggle with erroneous next action predictions. Fortunately, it is also the case that Precondition Verification works similarly well in both environments and is able to automatically satisfy the preconditions for erroneous action predictions at a high rate.

Overall, the detailed analysis in Tables \ref{tab:alfworld_detailed_experiments} and \ref{tab:ai2thor_detailed_experiments} demonstrate the ability of SCLPlan to effectively adapt its planning methodology to improve performance across different environments. This supports our initial hypothesis that symbolic planner augmentations can help improve the reliability and performance of LLM planners across environments of varying complexity. Our results also show that LLM usage tends to scale directly with environmental complexity.

\subsection{SCLPlan improves reliability and performance on real world systems}

We implement SCLPlan on a real-world robotic system using a Boston Dynamics Spot quadruped robot and benchmark its performance across 10 diverse, open-world tasks. Like previous experiments, we include ablations to demonstrate the effect of each symbolic planner augmentation. The results of the real-world experiments are shown in Table \ref{tab:real_world_experiments}.

\begin{table*}[htbp]
\centering
\caption{Real-world experiment results. All task plans executed on a real-world Boston Dynamics Spot quadruped.}
\begin{tabular}{p{7cm}cccc}
\textbf{Task} & \textbf{SCLPlan} & \textbf{ReAct + PV} & \textbf{ReAct Only} & \textbf{Symbolic Planner Only} \\
Place all discovered items in the red container & Success & Success & Failure & Success \\
Pick up and set down both objects & Success & Success & Failure & Failure \\
Move to all objects one by one & Success & Success & Success & Failure \\
Pick up the bag & Success & Success & Failure & Success \\
Place the apple in the red container and then look at the bag & Success & Success & Failure & Failure \\
Place the bag and the apple in the red container & Success & Failure & Success & Success \\
Bring the can to the traffic cone and drop it. & Success & Success & Failure & Failure \\
Place the apple in the red container and drop the bag beside the traffic cone. & Success & Failure & Success & Failure \\
Place the blue bag in the red container & Success & Failure & Success & Failure \\
I want you to place the blue bag next to the traffic cone. I hear the human knows something about the blue bag. & Success & Success & Success & Failure \\  &&&& \\
Overall Success Rate & 1.00 & .700 & .500 & .300 \\ &&&& \\
Percentage $ReAct$ Actions& 0.488 & 0.632 & 1.00 & - \\
Percentage $PV$ Actions & 0.262 & 0.368 & - & -   \\
Percentage $GSP$ Actions & 0.25 & - & - & 1.00 \\ \rule{0pt}{3ex}
\end{tabular}
\label{tab:real_world_experiments}
\end{table*}

We provide a list of each natural language task in the real-world task set and provide the success result of each individual task for all of our ablations. Additionally, we show the average success rate and provide data showing the proportion of actions generated by each component in the planning pipeline for all ablations across all tasks. In short, the results from our real-world experiments echo that of our simulation experiments: SCLPlan significantly improves planning performance over purely LLM and purely symbolic planning approaches. SCLPlan was able to successfully complete each of the 10 tasks in our real-world task set while the $ReAct + PV$, $ReAct$, and $Symbolic Planner$ were able to complete 70\%, 50\%, and 30\%, respectively. Similarly to the simulation experiments, SCLPlan is able to make full use of each of its planning components to improve performance with 48.8\%, 26.2\%, and 25\% of actions being predicted by the $ReAct$, $Precondition\ Verification$, and $Global\ Symbolic\ Planner$ components, respectively. Our $ReAct + PV$ ablation also supports the use of $Precondition\ Verification$, as 36.8 \% of predicated actions were generated by the $Precondition\ Verification$ component.

\section{Discussion}

The presented results demonstrate that pure LLM planners consistently fail when used on specialized systems, and further show that a hybrid planning approach can substantially improve the performance and reliability of AI planning systems. Furthermore, our research demonstrates that it is indeed possible to transfer these benefits from a simulation environment to a practical, real-world system to great effect. Our research demonstrates that a hybrid planning approach that blends LLMs with symbolic planners can adapt its planning methodology to meet planning needs in environments with varying complexity and significantly improves overall performance compared to baseline planning approaches across a variety of relevant metrics. We additionally demonstrate that our approach translates well when using a variety of LLM backbones highlighting its viability for smaller models running on the edge. Our approach is also compatible with future advances in dexterous manipulation and low-level control, as we focus mainly on the problem of high-level planning in which our planner is agnostic to the implementation details of its lower-level action functions. 

We envision a number of opportunities for future work. First, the real-world system presented in this work used brittle, model-based implementations for the low-level dexterous actions. This significantly limited the scope of possible tasks available to the agent. With the introduction of novel performant methods for low-level, dexterous manipulation the scope of possible tasks could increase significantly. The integration of such advances into our framework would be seamless, simply requiring the swapping of various functions for more performant versions. Second, after improving the low-level action implementations, our approach could be used to collect data for robotic foundation models. The reliability guarantees afforded by a sufficiently designed environment domain could allow for longer-duration task planning without the need for human intervention. The natural language reasoning traces provided by the LLM component of our planner could also provide data required for training a multi-modal model such as recent Vision Language Action models (VLAs). Finally, we explore only two symbolic planner augmentations in this work: 1) Precondition Verification and subsequent invalid state correction and 2) a global symbolic planner which attempts to solve the global task at each planning iteration. There are certainly more ways in which symbolic planners can be integrated effectively into an LLM planner pipeline in addition to the augmentations explored here. For example, you could break down a global task into smaller subtasks and attempt to solve each subtask with a global symbolic planner.

We also acknowledge some limitations of this work. First, there is currently a lack of high-quality planning benchmarks that truly test the reasoning capabilities of embodied systems. While ALFWorld and AI2Thor provide a decent starting point, the scope and difficulty of the provided tasks are limited in their complexity, making it difficult to exhaustively benchmark a planning approach with statistically rigorous means. Additionally, our real-world task set contains only 10 tasks due to the significant overhead required to set up each task in the physical environment, manually evaluate each plan rollout, and manage hardware usage and upkeep during experimentation. While we feel that our benchmarks are sufficient to demonstrate adequate real-world potential, subsequent research should work to introduce more statistically rigorous benchmarks that include larger, more diverse, and more complex task sets. We are hopeful that the community will overcome these benchmarking limitations in the near future.

In addition to the direct implications of our results, SCLPlan can serve as a broad framework for leveraging LLMs in practical systems. In practice, one loses nothing by leveraging our hybrid approach. If the environment is simple enough to fully constrain symbolically, you achieve optimal reliability and performance with our symbolic planning components. If the environment is too complex to fully define, at minimum you receive the baseline performance of an LLM planner with added reliability provided by our symbolic components that scales with the completeness of your environmental definition. Engineers and researchers can therefore thoughtfully decide on an acceptable trade-off between the engineering effort to fully define an environment and overall system reliability. We believe the presented results are both practical and relevant for researchers and engineers alike who require principled and performant ways of integrating LLMs into a general-purpose robotic system.

\section{Materials and Methods}

\subsection{Background and Problem Formulation}
In this work, we address the problem of closed-loop planning and execution of discrete action sequences guided by high-level, natural-language commands. This requires our agent to comprehend unconstrained and ambiguous natural language tasks, understand the physical environment in which it operates, and recognize its own capabilities to navigate and affect its environment to complete the high-level task. Formally, at each discrete planning timestep we provide our agent with a high-level natural language task, $\tau$, a set of $n$ actions $\mathcal{A}=\{\alpha_1, ..., \alpha_n\}$, a representation of the current environment state, $\mathcal{S}$, and a history of previously executed actions for the current task, $H = \{\eta_{0}, ..., \eta_{t-1}\}$, where $\eta = \{\alpha, \mathcal{O}\}$ is a tuple containing the action and corresponding observation feedback from the environment and $t$ is the timestep. SCLPlan, leveraging either a symbolic planner, $\theta_{SP}$, or an LLM planner, $\theta_{L}$, will predict the next action in the sequence, $\alpha_t$. Our agent will then execute this predicted action, receiving an updated $\{\mathcal{S}_t, \eta_t=\{\alpha_t, \mathcal{O}_t\}\}$ for the next iteration of planning. This loop continues until $\theta_{SP}$ or $\theta_{L}$ determines a goal state has been reached. 

\subsection{Defining an Environment Domain}
To utilize a symbolic planner, an environment domain must be defined. In this work, this process is done explicitly by a human expert, although other works have explored the use of LLMs for defining environment domains \cite{guan2023leveragingpretrainedlargelanguage, silver2023generalizedplanningpddldomains}.  Specifying an environment domain requires defining a set of $n$ predicates, $\mathcal{P}=\{p_1, ..., p_n\}$, that can be used to define attributes of entities (i.e. objects, locations, and the agent itself) in the environment. For example, \textit{isReceptacle} would be a valid predicate for an object of type \textit{Box}. Next, for each action we specify the natural-language action name, $\alpha_{name}$, natural-language description, $\alpha_{desc}$, and action arguments, $\alpha_{args}$. Using predicates taken from $\mathcal{P}$, we also specify a set of preconditions, $\mathcal{P}r=\{pr_1, ..., pr_n\}$, and effects, $E=\{e_1, ..., e_n\}$, for each action that defines the valid state required to enact an action as well as the effect that the action will have on the environment. In this way, one can symbolically constrain actions to be executable only when specific preconditions are satisfied. Additionally, this rigidly defines the outcome of the action within the environment in a way that is interpretable. This process is a standard requirement whenever using symbolic planners. We use PDDL \cite{pddl_original} to define our environment domain. 

\subsection{Planner Overview}

SCLPlan enables natural language-driven task planning and execution by harmonizing a formally defined symbolic planner with LLM reasoning to achieve improved reliability while maintaining impressive adaptability to complexity. Typically, a symbolic planner requires a \textit{complete} domain specification to handle all required tasks. Here we define a complete domain specification as one that defines a system for a given set of tasks such that each task in the set of tasks can be solved mathematically using search algorithms given infinite time and compute. For sufficiently complex environments and task sets, this is intractable. To remedy this, the introduction of our hybrid planning architecture allows users to specify only a \textit{partial} domain specification, as the adaptability of the LLM planner can account for uncertainty in the environment due to under specified domain definitions. In this way, a user must simply do the best they can instead of perfectly capturing an environment. A more complete domain specification will likely yield better results, but our framework provides the flexibility to bypass prior restrictions regarding symbolic planners. 

\subsection{Planner Architecture}
\begin{figure*}[htbp]
  \centering
  \includegraphics[width=\linewidth]{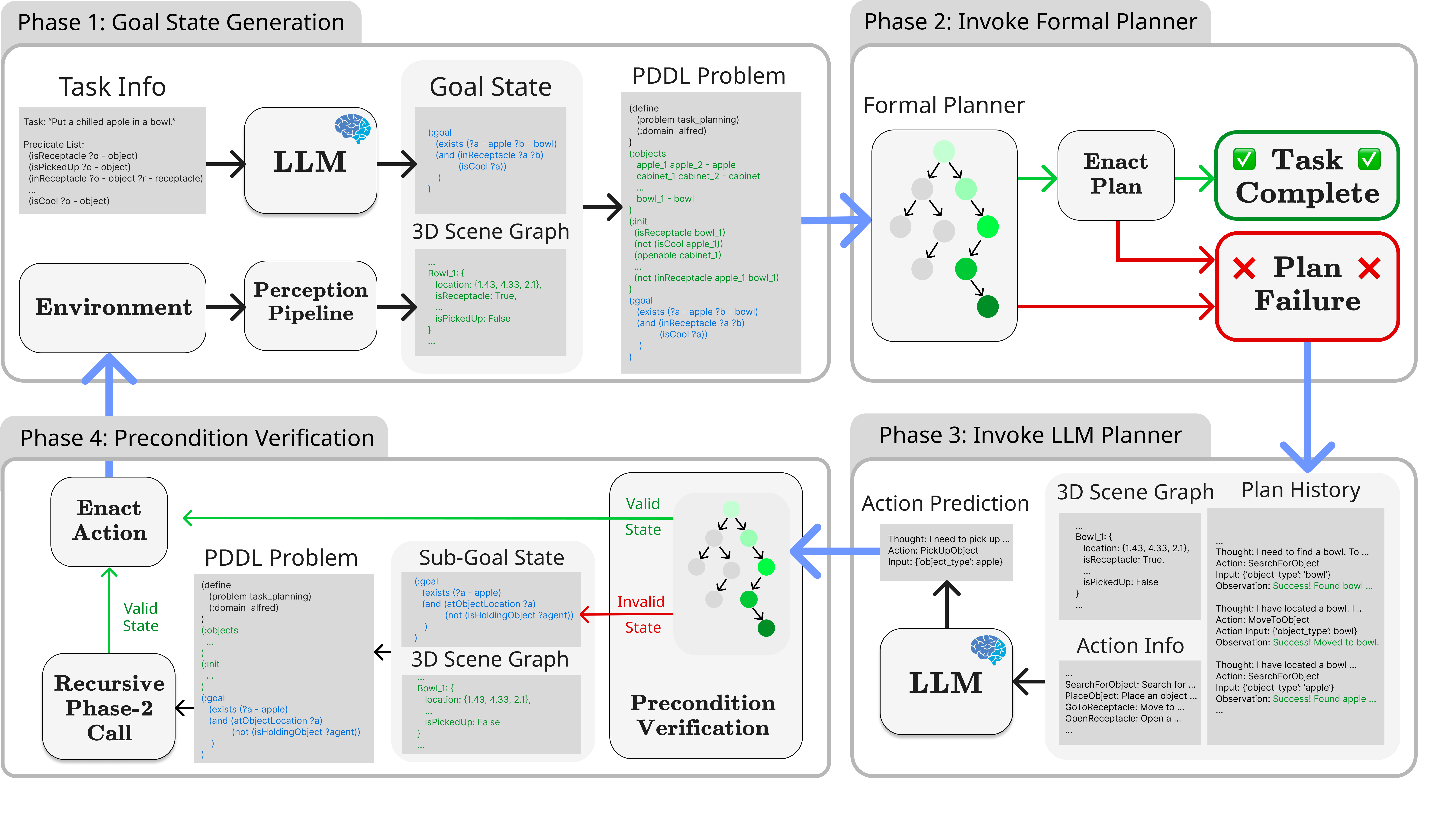}
  \caption{SCLPlan architecture.}
  \label{fig:FALA_architecture}
\end{figure*}

Our approach is modular, and the LLM, symbolic planner, and specified domain can be easily exchanged to accommodate model improvements, hardware requirements, different environments, and various robotic platforms. Figure \ref{fig:FALA_architecture} illustrates the full SCLPlan planning pipeline. SCLPlan is an iterative planner consisting of 4 phases. Each phase is ran sequentially at each planning iteration in a loop until the Task Complete flag is thrown. A detailed description of each phase is provided below.

\subsubsection{Goal State Generation}
In Phase 1, $\tau$ and $\mathcal{P}$ are passed as input to an LLM that is then prompted to define a Goal State, $\mathcal{G}$, in PDDL format. Our Perception Pipeline then creates a 3D Scene Graph that represents $\mathcal{S}_t$. For simulation experiments, the ground truth from the simulator environment was used. During our hardware experiments, a variety of off the shelf computer vision models and pointcloud projection techniques were used to build up a 3D scene graph in real time. $\mathcal{G}$ and $\mathcal{S}_t$ are then used to build a PDDL problem file that can then be passed to $\theta_{SP}$ for planning.

\subsubsection{Invoke Formal Planner}
In Phase 2, the PDDL problem file generated in Phase 1 is passed to $\theta_{SP}$ for planning. Although any heuristic search algorithm can be chosen, we use the BFS(f) solver \cite{lipovetzky2014width} implemented as an
executable by LAPKT \cite{ramirez2015lapkt} as our symbolic planner. At this stage, $\theta_{SP}$ attempts to take the LLM-defined goal state and the environment representation defined in the PDDL problem file and create an optimal plan that achieves that goal state. For tasks where a possible plan trajectory lies within the scope defined in the environment domain, $\theta_{SP}$ often produces an optimal plan quickly. If $\theta_{SP}$ generates a plan, each action in that plan is executed sequentially. If each action is executed successfully, the Task Complete flag is thrown and the planning and execution phase is considered complete. However, if an unexpected failure occurs during an action execution or if $\theta_{SP}$ cannot produce a valid plan to reach $\mathcal{G}$, we then pass a description of the failure onto Phase 3 of the pipeline.

\subsubsection{Invoke LLM Planner}
In Phase 3, $\theta_{SP}$ was unable to solve the planning problem or an unanticipated action execution error occurred. To dynamically adjust to planning failures and execution errors, we leverage the common sense reasoning capabilities of LLMs by using $\theta_{L}$ to intelligently predict the next action to recover from a prior planning failure. As SCLPlan is modular, any LLM agent can be chosen as the backbone for this phase. In this work, a Zero Shot ReAct \cite{react} agent is used. We choose a Zero Shot formulation as opposed to a Few Shot formulation for two reasons: $(i)$ we want to show that SCLPlan achieves impressive performance without requiring prompt engineering in the form of Few Shot examples for its LLM planner and $(ii)$ Few Shot examples muddle the evaluation process, as one could simply cheat the benchmark by tabulating out all possible tasks with corresponding plans in the few shot examples. To optimally leverage the `reasoning' capabilities of $\theta_{L}$, we provide as input natural language versions of $\mathcal{S}_t$, $\alpha_{name}$, $\alpha_{desc}$, $\alpha_{args}$, as well as a complete account of the planning and execution history of the current task. Since the Plan History will include the failure observation feedback from the previous planning iteration, this input provides all the context required to make an intelligent next action prediction. The $\theta_{L}$-predicted next action is then passed on to Phase 4 of the pipeline.

\subsubsection{Precondition Verification} In Phase 4, as LLMs are susceptible to hallucinations, SCLPlan improves upon the reliability of $\theta_{L}$ by verifying each predicted action through a novel Precondition Verification ($PV$) process. As the structured definition of the environment domain was required for $\theta_{SP}$, we can leverage this domain to achieve $PV$ without any extra overhead. Simply put, the preconditions for each action defined in the environment domain represent a state in which that action is considered valid. We can leverage $\theta_{SP}$ to determine if our environment is in a valid state with respect to the next predicted action by attempting to formally generate a plan with $\theta_{SP}$ to reach this valid state. If the generated plan has a total cost of 0 (i.e. no actions are required), then we are already in a valid state and the $\theta_{L}$-predicted action can be confidently executed. However, if the generated plan has an action sequence length greater than 1, we are not in a valid state for the $\theta_{L}$-predicted action. Fortunately, since we know the required state for the $\theta_{L}$-predicted action, we can formally define a Sub-Goal State in which the $\theta_{L}$-predicted action will be satisfied. Furthermore, we can optimally generate a plan to reach this valid environment state using $\theta_{SP}$ by recursively calling Phase 2 of the SCLPlan pipeline. In this way, we can achieve reliable next-action prediction from $\theta_{L}$ by leveraging $\theta_{SP}$ to greatly mitigate the effect of hallucinations. Once we reach the desired Sub-Goal State, our $\theta_{L}$-predicted action will be executed and the next planning iteration will begin at Phase 1 with an updated environment representation.

\subsection{Robot Setup}

In this work, we use a Boston Dynamics Spot quadruped with manipulator arm for all of our real-world experiments. We outfit our quadruped with a front-facing, binocular ZED camera which provides RGBD images of the robot's surroundings. We also leverage a manipulator arm camera that also provides RGBD images from the perspective of the manipulator arm. As we focus solely on planning performance and not on building a novel perception pipeline, we make various design choices with regard to our perception pipeline to achieve greater reliability. First, we define a set of relevant object classes with various static predicates (e.g. isReceptacle: `True', pickupable: `False', etc.). In the future, we envision the static predicate definition process could be performed by a trained model, although in this work this process was hand-performed by a human to ensure high reliability for accurately analyzing the downstream planning performance. To detect objects in the environment, we use the Grounding Dino \cite{liu2024grounding} open vocabulary object detector to detect each object class present in the list that was previously defined in the prior step. We then use the detected bounding boxes from Grounding Dino as prompts for the Fast Segment Anything Model \cite{zhao2023fast} to get a per-pixel segmentation of each class. We project this pixel segmentation into 3D space using the depth component of the RGBD image to achieve a point cloud representation of the objects. This affords our quadruped spatial knowledge of the environment. We leverage Boston Dynamics' off the shelf manipulator API for pick and place tasks and also leverage Boston Dynamics' off the shelf navigation software to navigate to and from objects and points of interest. As hardware experiments are prohibitively time-consuming, we test our performance only using GPT-4o as a backbone LLM in the real-world experiments.

\bibliographystyle{sciencemag}
\bibliography{bibliography}

\vfill

\end{document}